\documentclass[sn-apa,iicol]{sn-jnl}% Default with double column layout

\usepackage{subfig}
\usepackage[longnamesfirst,sort]{natbib}
\jyear{2022}%

\theoremstyle{thmstyleone}%
\theoremstyle{thmstyletwo}%

\theoremstyle{thmstylethree}%

\renewcommand{\vec}[1]{\mathbf{#1}}
\begin{document}

\title[Article Title]{Multi-view Tracking, Re-ID, and Social Network Analysis of a Flock of Visually Similar Birds in an Outdoor Aviary}


\author[1]{\fnm{Shiting} \sur{Xiao}}\email{gxiao@seas.upenn.edu}
\equalcont{These authors contributed equally to this work.}

\author[1]{\fnm{Yufu} \sur{Wang}}\email{yufu@seas.upenn.edu}
\equalcont{These authors contributed equally to this work.}

\author[2]{\fnm{Ammon} \sur{Perkes}}\email{aperkes@sas.upenn.edu }
%\presentaddresstxt{Department of Evolution and Ecology, University of California, Davis, Davis, CA, USA}

\author[1]{\fnm{Bernd} \sur{Pfrommer}}\email{pfrommer@seas.upenn.edu }

\author[2]{\fnm{Marc} \sur{Schmidt}}\email{marcschm@sas.upenn.edu}

\author[1]{\fnm{Kostas} \sur{Daniilidis}}\email{kostas@cis.upenn.edu}

\author*[1]{\fnm{Marc} \sur{Badger}}\email{mbadger@seas.upenn.edu}

\affil[1]{\orgdiv{Department of Computer and Information Science}, \orgname{University of Pennsylvania}, \orgaddress{\city{Philadelphia}, \state{PA}, \country{USA}}}

\affil[2]{\orgdiv{Department of Biology}, \orgname{University of Pennsylvania}, \orgaddress{\city{Philadelphia}, \state{PA}, \country{USA}}}


% Statement to introduce topic and hook reader
% Statement about why this topic is impactful and is the problem
% What we do to address the problem (contributions).
% List one or two main results.

% Check < 200 words
\abstract{The ability to capture detailed interactions among individuals in a social group is foundational to our study of animal behavior and neuroscience. Recent advances in deep learning and computer vision are driving rapid progress in methods that can record the actions and interactions of multiple individuals simultaneously. Many social species, such as birds, however, live deeply embedded in a three-dimensional world. This world introduces additional perceptual challenges such as occlusions, orientation-dependent appearance, large variation in apparent size, and poor sensor coverage for 3D reconstruction, that are not encountered by applications studying animals that move and interact only on 2D planes. Here we introduce a system for studying the behavioral dynamics of a group of songbirds as they move throughout a 3D aviary. We study the complexities that arise when tracking a group of closely interacting animals in three dimensions and introduce a novel dataset for evaluating multi-view trackers. Finally, we analyze captured ethogram data and demonstrate that social context affects the distribution of sequential interactions between birds in the aviary.}




\maketitle

\section{Introduction}\label{introduction}

% Why study interactions between individuals?
In social animals, moment-to-moment interactions among individuals drive the formation of long-term social networks. In turn, both an animal’s position in the social network and its immediate social context change how it behaves and interacts with others \citep[e.g.][]{White2010, Anderson2021}. The dynamics of a group’s social network drives how individuals access food, shelter, and mates, and ultimately determines the group’s reproductive success \citep{Kohn2013}. As we work toward a quantitative understanding of social behavior, it is essential that we develop animal and engineering systems for studying the interplay between the behavior of individuals and group dynamics.

% Capturing social behavior is hard
Capturing the dynamics of social networks is not an easy task. Individuals must be accurately tracked and re-identified over long time periods and interactions between individuals must be detected and characterized to create an ethogram, or record of salient behaviors and their timestamps, for all individuals. Manual focal sampling by behavioral experts is one way of creating ethograms, but such efforts only capture a small slice of important behaviors for a few individuals at a time. Many recent works have developed automated systems supporting the creation of behavioral ethograms, including those focusing on 2D tracking and re-ID \citep{idtracker, idtrackerai, trex}, pose estimation in 2D \citep{Mathis2018, Lauer2022, mars, Pereira2019, Pereira2022sleap, DeepPoseKit, AlphaTracker}, and 3D \citep{LiftPose3D, OpenMonkeyStudio, ActinoSet, Dunn2021, DeepFly3D, badger2020, wang21aves, zuffi2019three}, behavioral mapping \citep{Berman2014}, and analysis of collective behavior \citep{Heras2019, Katz2011, Evangelista2017}.

% What problems challenge the current state of the art?
Of foundational importance to all multi-animal pipelines is the ability to track and re-identify individuals. With a few exceptions \citep{DeepPoseKit, ActinoSet, badger2020}, current systems have only been deployed and tested in 2D settings with consistent lighting and static backgrounds, which make the problems of detection and tracking significantly easier. Interesting social dynamics, however, usually do not occur in isolation. Instead, they are embedded in the surrounding 3D environment, which introduces many challenges for automated perception. Groups of interacting animals spread over regions orders of magnitude larger than their body size, requiring many cameras to capture details for every individual. Individuals may be visually similar, yet their appearance may change dramatically as they move in 3D, puff their fur, or fluff their feathers. Variable lighting further alters the appearance of individuals. Backgrounds are visually complex and dynamic, and animals are frequently occluded by each other and structures in the environment. Many animals also have multimodal motion distributions making tracking extremely difficult. The extent to which automated systems can overcome these difficulties and capture groups of animals interacting within large and complex 3D environments is not well understood.

% Details of what we are presenting (briefly how we tackle the multi-view multi-animal tracking problem in our challenging setting, and introduce our dataset).
In this work, we aim to study behavioral dynamics in a socially gregarious species of songbirds \citep{White2012, Maguire2013sociallesion}. We present 1) approaches for tracking a flock of birds and capturing their social interactions in a dynamic, multi-view setting, and 2) a new challenging dataset for evaluating the real-world performance of multi-view multi-object trackers. 
 
Tracking in 3D is a complex problem. Some methods perform 3D reconstruction followed by tracking (Reconstruction-then-Tracking, or RT) and other methods first form tracks in 2D and then associate the tracks across views (Tracking-then-Reconstruction, or TR) \citep{wu2009}. The advantage of performing reconstruction first is that tracking ambiguities are much less common in 3D than in 2D, so associating detections across time is far easier in 3D. On the other hand, matching sequences of points from 2D tracks improves cross-view association by reducing the potential for false matches, which create ghost trajectories. When used for tracking bats, these two approaches show a tradeoff between the number of track fragments and false positive tracks \citep{wu2009} and the best-performing approach will depend on both camera geometry and the performance of the 2D tracker. We implement two RT approaches because the camera views frequently contain many occlusions and the baseline 2D trackers such as SORT \citep{Bewley2016_sort} did not perform well under these situations.

Our first approach uses foreground masks to construct a 3D pointcloud, which is then clustered to form points for tracking in 3D. Our second approach performs stereo matching of detections across views to reconstruct 3D points. In both approaches, 3D points are subsequently linked over time to form tracks using a motion prior. We test the performance of both trackers on an evaluation dataset containing long trajectories ($\sim$36000 frames) with sparse 3D annotations and ground truth identities.

Our evaluation dataset includes a challenge task along with code for loading and viewing examples and evaluating performance on the task. In the task, which we call Where’d It LanD (WILD), the 3D locations of a single bird’s head and tail are provided along with a sequence of frames. The tracker must then return the 3D location of the same bird's head at the end of the sequence as the target bird hops or flies with other birds in the aviary. Predictions are marked as correct if the returned 3D location is within a given threshold distance of the ground truth 3D location. Tracking performance is evaluated by the fraction of correctly predicted sequences across a range of distance thresholds.
Finally, we use our dataset to perform a behavioral analysis of birds interacting in the aviary and show that social context influences the distribution of actions used by birds during courtship.

\begin{figure*}[hbt!]
\begin{center}
\includegraphics[width=\linewidth]{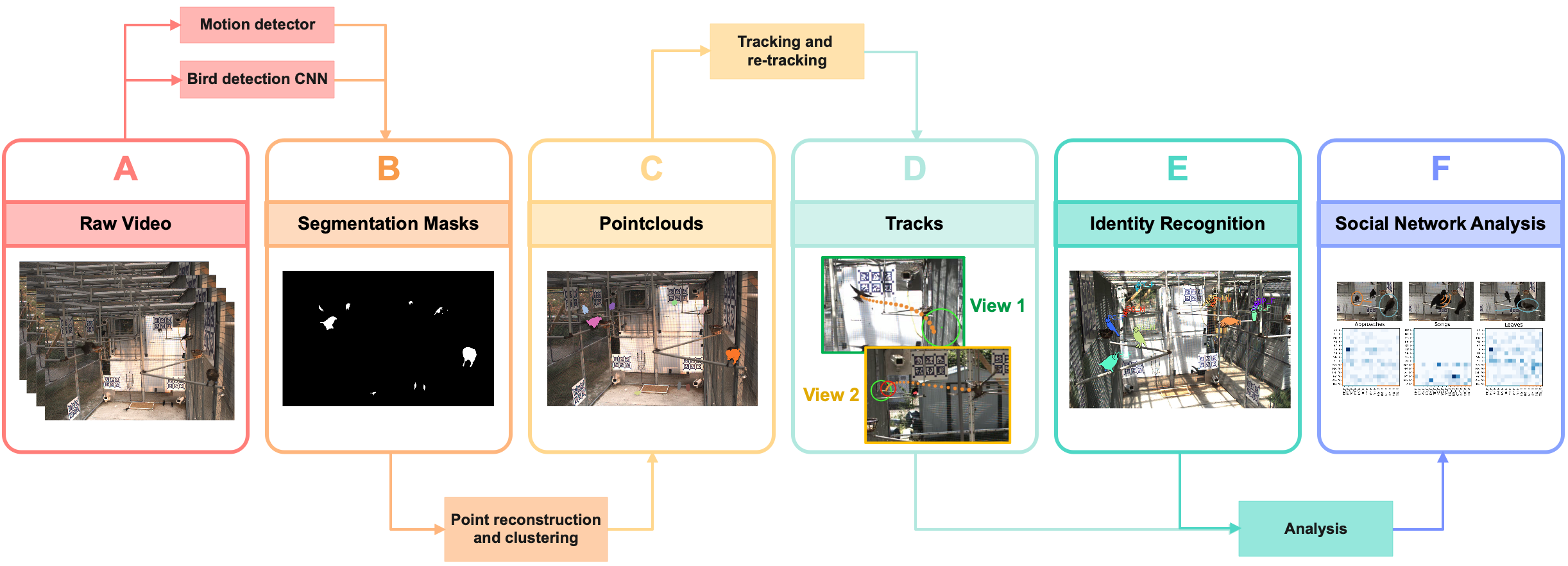}
\end{center}
\caption{\textbf{Full pipeline for cowbird tracking and recognition.} (A) A synchronized set of raw videos from multiple views are processed in a frame-by-frame manner. (B) Segmentation masks of bird instances are obtained using a Mask R-CNN network and background subtraction. (C) Pointclouds are reconstructed by multi-view matching, triangulation, and clustering. (D) Tracking, which is implemented using a Lagrangian Particle Tracking (LPT) algorithm, links pointclouds in time to form tracklets. Re-tracking associate 3D tracklets to generate longer 3D tracks. (E) Individual identity recognition using the FastReID framework. (F) Output from the pipeline can then be used for social network analysis. }
\label{fig:tracking_pipeline}
\end{figure*}

\section{Contributions}
\begin{enumerate}
\item A system for automatically extracting behavioral ethograms from a flock of birds interacting in an outdoor aviary. Components include synchronized camera and microphone array recording for months-long durations, and pipelines for detection, reconstruction, tracking, and re-identification.
\item An exploration of reconstruction-then-tracking approaches to multi-view multi-object tracking.
\item A unique dataset and codebase with tracking challenges for evaluating multi-view multi-object tracking algorithms.
\item An analysis of the social network of a flock of cowbirds showing how social context affects behavioral choices made by male and female birds during courtship.
\end{enumerate}

\section{Related work}\label{related_work}
\subsection{Multi-object tracking}
\subsubsection{Detection}
Most state-of-the-art tracking methods follow the tracking-by-detection paradigm \citep{bergmann2019, bewley2016simple, karunasekera2019multiple, wojke2017simple, wu2009, cavagna2021Sparta, sinhuber2019insect, ling2018stereomatching}, in which the quality of detection is critical to the tracking performance.  Convolutional Neural Network (CNN) based detectors \citep{girshick2014rich, girshick2015fast, he2017mask, ren2015faster, liu2016ssd, redmon2016yolo, wang2020solo, lin2017focal} have outperformed previous methods for object detection and instance segmentation tasks. 
In particular, the R-CNN family \citep{girshick2014rich, girshick2015fast, he2017mask, ren2015faster} find category-agnostic bounding box candidates, and then perform classification and refinement on them based on feature maps. 
A latest work Context R-CNN \citep{beery2020context} keeps a ``memory bank'' based on contextual frames and uses attention to improve detection. 
SSD \citep{liu2016ssd}, the YOLO family  \citep{redmon2016yolo, wang2020solo}, and RetinaNet \citep{lin2017focal} directly regress to category-specific bounding box candidates. 
Detection can fail though, if an object's appearance changes dramatically between sightings. Unless enough examples are available in the training data, networks may not be robust to such changes. In the aviary, for example, motion blur caused by birds in flight is rare in training data and hence difficult to detect.
Background subtraction is a widely used technique to detect dynamically moving objects from static cameras.
\citet{zivkovic2004} and \citet{zivkovic2006} use a Gaussian mixture model that captures gradual changes in the background such as illumination changes, which is an important factor when running outdoor experiments where the sun is the light source. 
By using both a CNN based detector and a background subtraction based motion detector, we can reliably detect birds despite variations in their postures and movements.

\subsubsection{Trajectory Generation}
The ability to track an individual animal as it moves throughout its 3D environment is fundamental for addressing a broad range of questions in behavioural ecology and the study of animal social networks.
Some interesting methods obtain 3D detections using point cloud observations from LiDAR data \citep{Weng2020_AB3DMOT, chiu2020probabilistic, yin2021center}, but obtaining such data is unrealistic in long-term wildlife monitoring. 
Recently, video data has become ubiquitous and indispensable in the study of collective behavior \citep{caravaggi2017review, schofield2019chimpanzee, ling2018stereomatching, sinhuber2019insect}.
When individuals interacting in a 3D environment pass behind each other or objects in the environment, 2D occlusions occur. Because single camera views do not provide depth information, such occlusions create ambiguities and often result in lost tracks, identity swaps, or other tracking errors \citep{ciaparrone202061}. Occlusions occur more frequently in crowded environments and identity swaps that occur during such occlusions can be difficult to recover from if animals have similar appearances. 
An intuitive solution is to use multiple calibrated cameras and fuse information from different viewpoints to resolve ambiguities.

To track multiple objects in multiple camera views, data association must be performed not only across time (Tracking), but also spatially across views (Reconstruction). Doing reconstruction and tracking at the same time is computationally infeasible \citep{atanasov2014semantic}, so current methods typically adopt either a Tracking-then-Reconstruction (TR) route or a Reconstruction-then-Tracking (RT) route \citep{wu2009, cavagna2021Sparta}. 
TR methods first form 2D tracks in each camera views and then match them to reconstruct 3D tracks. 
Many state-of-the-art 2D tracking algorithms \citep{bergmann2019, bewley2016simple, karunasekera2019multiple, wojke2017simple} can be readily extended to track in 3D using cross-view data association techniques \citep{wang2014tracklet, wu2016global}, but the complexity of most data association methods grows quickly with the number of simultaneously processed frames. 
Working in the 2D space, TR methods also have to handle both 2D and 3D occlusions in the reconstruction procedure \citep{cavagna2021Sparta, wu2016global}.

Conversely, RT methods first reconstruct 3D representations using cross-view matching techniques, and then link them in time to form 3D trajectories. 
2D occlusions are solved during the reconstruction procedure, which is typically performed independently for each frame, so the complexity of RT methods is substantially lower than the TR methods.
\citet{sinhuber2019insect, ling2018stereomatching} associate detections from multiple camera views using the stereo matching technique and use predictive Lagrangian Particle Tracking (LPT) \citep{ouellette2006quantitative} to form short 3D trajectories, or tracklets. A re-tracking strategy \citep{xu2008tracking} is then used to solve 3D occlusions and link these short tracklets to form longer trajectories.
A recent RT work by \citet{cavagna2021Sparta} reconstructs each target as a point cloud in 3D and resolves 3D occlusions by solving a partitioning problem through a semi-definite optimization technique.
While this method has proven to be effective for tracking birds moving at non-zero velocities in a dense flock, it performs poorly and cannot separate birds that perch close together for minutes (several hundred frames) because the complexity of the partitioning problem becomes too high to be solved reliably.
Beyond using simple 2D locations to reconstruct 3D representations of targets, other methods also encompass orientation \citep{Cheng2015fruitfly}, keypoints \citep{dong2021fast}, and deep appearance features \citep{dong2021fast, zhou2015multi} to perform association across views.
In this work, we only use 2D locations and masks to reconstruct the targets in 3D for simplicity and efficiency.

\subsubsection{Datasets}
State-of-the-art multi-object tracking (MOT) datasets predominantly target people and vehicles, motivated by surveillance and self-driving applications \citep{sun2020scalability, gan2021mvmhat, han2021mmptrack}. Datasets for animal tracking and related tasks are presented by a comparatively small amount of previous literature. 
Recent work AP-10K dataset \citep{yu2021ap} is the first large-scale benchmark for mammal animal pose estimation which consists of 10,015 images from 23 animal families and 54 species. The OVIS dataset \citep{qi2021occluded} for video instance segmentation consists of 20 animal species in hundreds of occluded scenes. Recently, a larger scale dataset for Tracking Any Object (TAO) \citep{dave2020tao} has been compiled containing 2,907 videos.
We contribute our multi-view 3D tracking dataset of cowbirds for evaluating generalist trackers.

In the biology context, most behavioral studies acquire the dataset with carefully designed lab conditions: ideal illumination, arenas with a plain background, and well-quantified or no environmental stimuli \citep{sinhuber2019insect, idtracker, idtrackerai}. While well-defined lab environments make it easier for tracking the objects, they restrict the complexity of the objects’ movements that can be measured. Birds, in particular, exhibits rich postures and movements. Current datasets for the tracking of birds, however, contain only scenarios of bird flocks in migration \cite{ling2018stereomatching, wu2014thermal}. In contrast, our multi-view tracking dataset contains large variation in bird pose, orientation, appearance, and social interaction across different lighting conditions that characterize “wild” footage.

\subsection{Animal Re-Identification}
 In spite of the vast literature on multi-object tracking, handling occlusions remains the biggest challenge, especially in crowded scenes. Visual appearance features can aid frame-to-frame association \citep{DeepSORT, idtrackerai, Pereira2022sleap}, and the ability to re-identify (re-ID) an individual animal upon re-encounter is extremely helpful in preserving the correct identities after occlusions.
 However, few ecological studies have taken advantages of the deep learning re-ID methods despite their success in human re-ID \citep{schneider2018}. 
 More recently, \citet{schofield2019chimpanzee} used a variant of the VGG-M architecture \citep{chatfield2014return} for both identity and sex classification of wild chimpanzees. When pre-trained on the ImageNet dataset, the VGG19 CNN architecture \citep{simonyan2014very} can recognize individuals within small groups of birds \citep{ferreira2020deep} and giant pandas \citep{hou2020panda}.
 While classification approaches have demonstrated good overall performance \citep{Luo_2019_CVPR_Workshops} and can generalize across age-related changes in individual appearance \citep{schofield2019chimpanzee}, the extent of their generalizability to unseen individuals in a small dataset (small in the number of individuals and training examples) is an important question that remains unexplored.
 Deep metric learning approaches, on the other hand, have shown good generalization across difference individuals and datasets \citep{yi2014deep, zou2021person}. Here we collect a dataset for bird re-identification and train an identity embedding network using both metric-learning-based and classification-based losses \citep{Luo_2019_reid}.

\section{Data collection}\label{methods}

\subsection{Aviary}
Many songbird species exhibit complex social structures, including the highly gregarious brown-headed cowbirds (\textit{Molothrus ater}). Cowbirds present an excellent study system because exhibit complex patterns of behavioral interactions and the dynamics and structure of a group’s social network predicts  overall reproductive success \citep{Kohn2013}. Interactions between birds occur on timescales ranging from seconds to months. In just a few seconds a male could sing aggressively towards another male and then fly toward and land near a female, who then might make a chatter vocalization, lunge at the male, or fly away. Through hundreds of these interactions pair bonds between males and females emerge and a stable social network forms over the course of the three month breeding season. Several interesting questions remain unanswered, including what interactions influence the formation of pair bonds between males and females, how these interactions change over time, and how female feedback and multi-way interactions influence the development of the social network throughout the breeding season. Furthermore, these dynamics and the possible quantification of the social network will allow for eventual neurobiological studies that probe the influence of social context on brain dynamics in a naturalistic environment. To address these questions, we studied a flock of 15 cowbirds housed in a large outdoor aviary.

The UPenn Aviary is a covered outdoor arena (length $\times$ width $\times$ height: 6 $\times$ 2.4 $\times$ 2.4 meters) enclosed by rigid wire mesh. Inside are 12 central perches (located 40 cm below the ceiling) and 8 additional perches on the long sides (50 cm below the ceiling) of the aviary (see Figure \ref{fig:tracking_results}b,c for a diagram). Each corner has one camera (BLFY-PGE-23S6C with a Kowa 12.5 mm C-Mount lens) pointing inwards. The height $\times$ width field of view of the cameras is approximately 31 $\times$ 48 degrees and they are angled so that all points in the aviary volume can be observed by at least two cameras. Ten of the twelve central perches can be seen by all four top cameras. The bottom four cameras capture birds when they descend the ground to feed or bathe. Cameras are synchronized by a hardware trigger and capture 1920 $\times$ 1200 pixel frames at 40 Hz, which are sent over Gigabit Ethernet to a central server. Cameras are calibrated using a standard checkerboard (intrinsics) and an array of 96 AprilTags \citep{krogius2019iros} printed on 16 aluminum boards attached to the walls of the aviary (extrinsics). The aviary also captures audio signals using an array of 24 microphones (Behringer ECM8000), which are organized in eight triplets (with $\sim$ 10 cm between microphones within a triplet) around the exterior of the aviary and sampled at 48 kHz. The server writes all camera and microphone messages and their timestamps to one ROS bag \citep{Quigley09} for each day of recording.

\begin{figure}[hbt!]
\begin{center}
\includegraphics[width=0.98\linewidth]{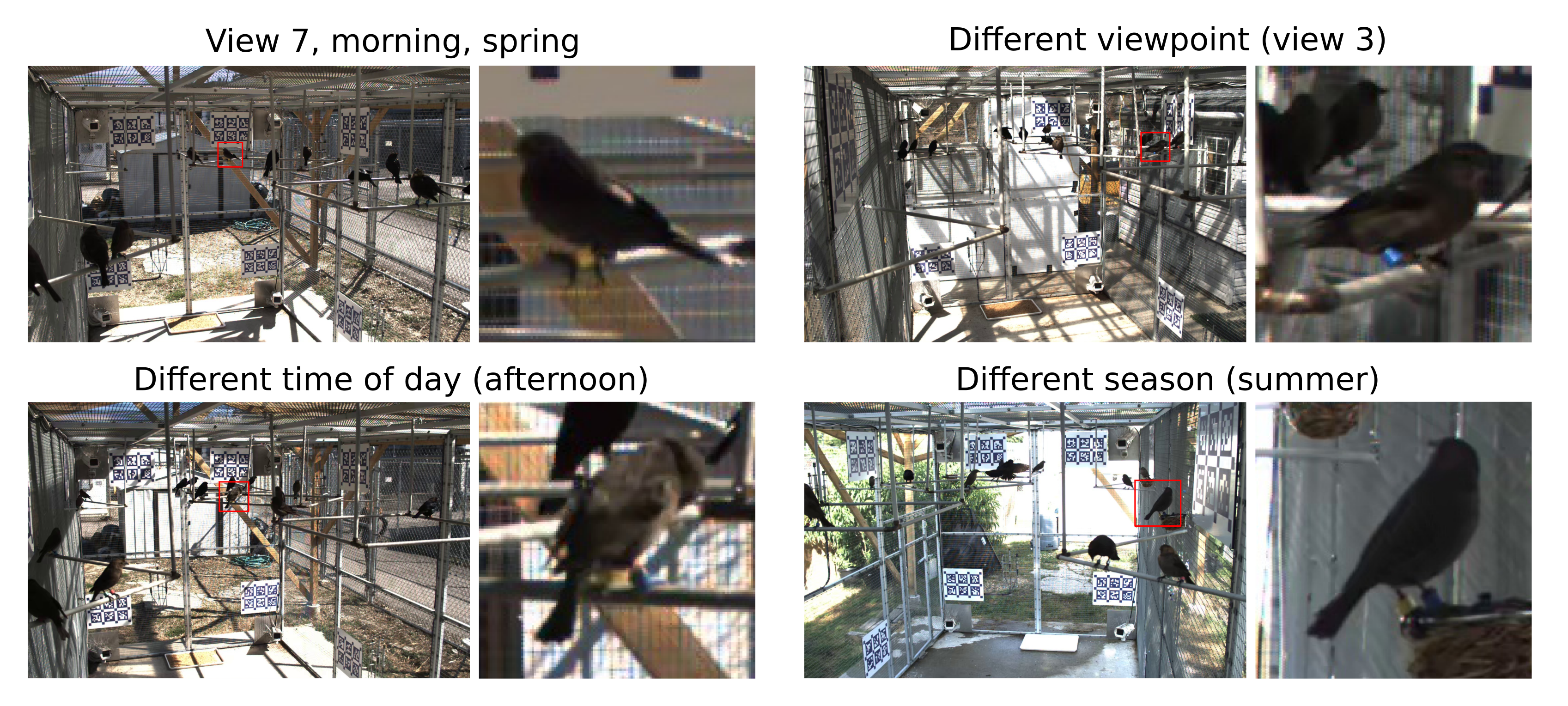}
\end{center}
\caption{\textbf{Variation of captured images.} Lighting and background appearance varies widely across viewpoint, time of day, and season throughout the birds' breeding period.}
\label{fig:aviaryvariation}
\end{figure}

Using the recording system described above, we recorded a flock of 15 interacting cowbirds (\textit{Molothrus ater}) for approximately 16 hours per day for 104 days (March 16, 2019 to June 28, 2019). Captured images varied significantly in appearance across views and with the time of day, weather, and season (Figure \ref{fig:aviaryvariation}). There were six males and nine females in the flock. Males have black bodies with dark brown plumage on their heads and are larger than females, which are brown colored with lighter gray-brown breasts (see Figure \ref{fig:tracking_results}a for examples). We banded the left and right legs of each bird with a unique color combination drawn from blue, teal, green, pink, red, and yellow colors. Leg bands were approximately 1 cm in diameter and birds could be manually identified from nearby cameras whenever there was sufficient lighting and their bands were not occluded. Birds usually perched on the perches when not flying around the aviary, but they occasionally perched on the walls or walked along the floor between food and water trays. Perching periods varied dramatically, lasting from a fraction of a second to over 15 minutes. During long periods of perching, shadows shifted more rapidly than the birds themselves. In flight, however, birds crossed the 6 meter aviary in about 1 second (40 frames) and moved more than a body length between consecutive frames.

\subsection{Multi-view multi-bird dataset and challenge tasks}
Our dataset for multi-view multi-object tracking originates from four 15 minute segments drawn from one day in early April and two days in mid May. We chose these months because we expected to see rapid change in the social network across this period. The social network, including pair bonds, is not yet formed in April but solidifies by mid-May. Because cowbirds’ behavior in the aviary makes it relatively easy to annotate periods of perching, we chose to annotate the beginning and end of these stationary periods for every bird in the aviary. 

Each annotation effort began by selecting a bird and viewing a synchronized multi-view recording from the aviary in the VIA Video Annotator \citep{dutta2019vgg}. Once a bird stopped flying or walking (e.g. by landing on a perch), the center of the bird’s head and the tip of its tail were clicked in at least two views. Very small motions during stationary periods (\textless 10 cm), such as steps along the same perch, were annotated with midpoints. Just before the bird started its next flight, its head and tail were annotated and labeled as an end point of the stationary sequence. The bird was then followed visually in flight until it landed again and a new stationary sequence was started. A behavioral annotation was also created whenever a target male sang.
We ignore female chatter vocalizations because the visual chattering cue is subtle and annotators had a hard time assigning chatter when the female was not close to the camera. We plan to incorporate sound detection and localization to reliably assign chatters in future work.
We confirmed the identity of each bird whenever both its leg bands were visible. All 15 birds in all four segments were positively identified and no two birds in the same segment were given the same identity. After all birds were annotated for a given segment, annotations were triangulated to obtain a sparse sequence of 3D locations and body axis orientations for each bird. For stationary segments, the positions of the head and tail were interpolated between the start and endpoints (using any available midpoints). 
Annotations were inspected for tracking errors (ID swaps or merges) by plotting pairwise distances between all birds. Whenever the distance between any two birds became less than 15 cm, the annotations were manually checked to ensure that trajectories had not merged (i.e. that no identity merge had occurred during manual annotation). From the annotations, we extracted 1098 stationary sequences of widely varying length. Averaged across birds, the 10th, 50th, and 90th percentiles of stationary sequence length were 3.7, 17.6, and 165 seconds respectively. These stationary sequences were used to form a training dataset for re-ID described below.

\begin{figure}[hbt!]
\begin{center}
\includegraphics[width=0.98\linewidth]{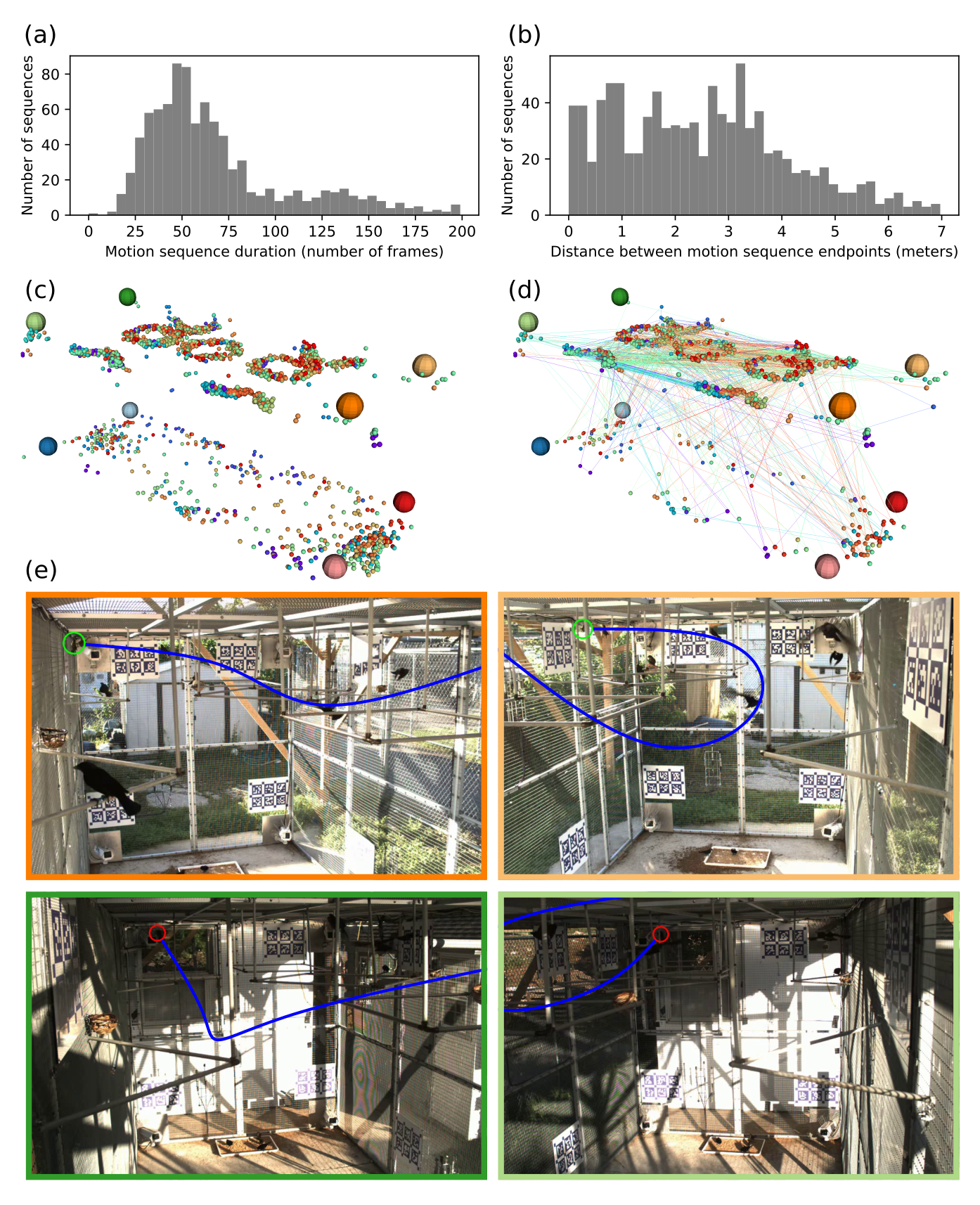}
\end{center}
\caption{\textbf{The WILD dataset.} Motion sequences are usually between 15 and 200 frames (a) between endpoints separated by 0-6 meters (b). In a reconstruction of all stationary sequence start and end points (c), areas of high point density reveal the perch geometry and ground plane. Lines between motion sequence start and end points (d) reveal flights from perch to perch, and from perches and the ground. Lines connect start and end points belonging to the same sequence; they do not indicate the actual trajectories. Points in (c) and (d) are colored by bird ID. Large spheres show the locations of the camera centers. An example from the dataset (e) shows the target bird's start location (green), approximate flight trajectory (blue), and ending location (red). Image borders denote the camera and correspond to large sphere colors in (c,d).}
\label{fig:wild_dataset}
\end{figure}

Untracked periods between stationary sequences were collected to obtain 986 motion sequences and formed our “Where’d It LanD” or WILD challenge. Each motion sequence is annotated with 3D start and end points (Figure \ref{fig:wild_dataset}e; the endpoint of a stationary sequence serves as the start point of the following motion sequence). Averaged across birds, the 10th, 50th, and 90th percentiles of motion sequence length were 0.88, 1.6 and 4.5 seconds (35, 63, and 180 frames) respectively (Figure \ref{fig:wild_dataset}a). The average number of motion sequences per bird was 66 (minimum: 8, maximum: 269) or an average total duration of 157 seconds per bird (minimum: 15.5 s, maximum: 552 s). The mean distance between motion sequence endpoints was 1.9 m (Figure \ref{fig:wild_dataset}b, d).

Motion sequences in WILD vary dramatically in difficulty. In “easy” examples, a bird might hop between two perches and the entire sequence can be seen from the same set of cameras (e.g. Figure \ref{fig:tracking_results}b). In more challenging examples, birds change direction multiple times, fly behind other birds or through dark areas, or land in areas that are not visible by the original set of cameras (e.g. Figure \ref{fig:tracking_results}c). In the most difficult cases, birds might be fully visible by only one camera and be partially or fully occluded from view by a second camera, and might then fly and land in an opposite corner of the aviary, where they are not visible by the original set of cameras (e.g Figure \ref{fig:wild_dataset}e).

As part of the WILD challenge, we provide a data loader that takes in an example index and returns metadata, 3D start and end points of the target bird and an iterator containing the sequence of synchronized multi-view frames. We also provide an example visualization script that creates a video showing the start and end points of a sequence reprojected onto all visible views. Finally, we provide an evaluation script that takes in a list of indices and predicted 3D endpoint locations and returns the fraction of correctly predicted sequences using several distance thresholds.

\section{Multi-view multi-bird tracking}
\subsection{Approach}
We present an automated pipeline that can detect and track multiple cowbirds from raw video footage and demonstrate its use on the WILD challenge. 
The pipeline consists of the following components: 
(A) detection of bird instances using a combination of a Mask R-CNN \citep{mask-rcnn} network and a Gaussian Mixture-based background subtraction algorithm \citep{zivkovic2004},
(B) multi-view reconstruction of 3D points in a frame-by-frame manner,
(C) 3D tracklets generation using a predictive Lagrangian Particle Tracking (LPT) algorithm,
and (D) occlusion handling in a re-tracking procedure.

\subsection{Detection}
We use a Mask R-CNN network pretrained on COCO instance segmentation to localize bird instances. 
Similar to our previous work \citep{badger2020}, we removed weights for non-bird classes (leaving bird and background) and then fine-tuned all layers on on the Aviary Dataset \citep{badger2020}. 
While Mask R-CNN would be robust to variations of bird postures given enough training examples, it is not reliable when detecting birds in certain postures which are rarely seen in training data, such as birds in flight with motion blur. 
To account for this issue, we add a background subtraction module \citep{zivkovic2004} to detect flying birds.
For each frame in a raw video, we first convert it to a grey scale image, and then remove stationary features from the scene, eg. the aviary settings and gradual changes in illumination, adaptively learned from 500 temporally consecutive frames using Gaussian mixture probability density. 
We then segment the foreground image into distinct blobs of pixels corresponding to bird instances. 
However, shadows often move faster than perched birds, so pure background subtraction is not reliable when capturing birds that remain stationary during a substantial part of the video footage. 
We therefore exploit advantages of both Mask R-CNN detector and motion-based detector, keeping a union of their detections without duplicates as inputs to the next stage of the pipeline.
By combining the two methods, we are able to reliably detect birds both in stationary and in motion.

\subsection{Reconstruction}
We use a similar method to \citep{cavagna2021Sparta} to reconstruct points in 3D.
At each instant of time, given a union of segmentation masks from each camera view, we find matched pairs of active pixels from 2 distinct camera views based on epipolar distance. 
In the aviary, a region can be seen in another 2-4 camera views. 
We consider a pair to be a good match if it satisfies the trifocal constraint \citep{hartley2003multiple} with another active pixel from one of those views. 
The matched pairs of pixels are then triangulated using a standard DLT method \citep{hartley2003multiple}.
A potential challenge may occur if a bird were to enter the camera view at an extremely near distance, which results in a big mask with a large number of pixels that could blow up the memory. To solve this, one could  sub-sample a mask if the number of pixels in it exceeds certain number.

After reconstructing all 3D points, ghost points due to bad triangulation or false detection are filtered temporally if their nearest neighbor cannot be found in the neighboring frames. We then cluster the 3D point clouds using the DBSCAN clustering algorithm. Centers of the clusters are the inputs to the tracking algorithm described in the next subsection.

\subsection{Tracking}
Once the 3D positions of the detected bird instances are reconstructed at each instant of time, we link them in time through an LPT \citep{ouellette2006quantitative} procedure. 
This tracking method has been successfully applied to study dynamic behaviour in aggregations of animals, including swarms of midges \citep{sinhuber2019insect} and flocks of birds \citep{ling2018stereomatching}. 

At a generic time $t$, let $\vec{x}^t_i$ denote the $i$th 3D point. The objective of the tracking problem is to find an $\vec{x}^{t+1}_j$ for every $\vec{x}^t_i$ such that $\vec{x}^{t+1}_j$ corresponds to the 3D location of the point at time $t+1$ that was at position $\vec{x}^t_i$ at time $t$. 
We define $\phi^n_{ij}$ to be the cost of associating each pair of $\vec{x}^t_i$ and $\vec{x}^{t+1}_j$. 
As this multidimensional assignment problem and is known to be NP-hard \citep{ouellette2006quantitative}, minimizing the overall cost spanning hundreds of frames is computationally expensive. 
Therefore, we limit the temporal association to only a few frames at a time.

We generate 3D trajectories for each individual in the following two stages:
\begin{enumerate}
    \item \textit{Tracking}: Associate 3D points in time to form short tracklets in a frame-by-frame manner.
    At first instant of time, $t = 1$, we perform Hungarian matching based only on the distance between points as there's no dynamic information from the past. For each matched pair of points, we add a velocity vector to points at $t = 2$ defined as follows:
    \begin{equation}
        \vec{v}_j^2 = \frac{1}{\Delta t} (\vec{x}^{2}_j - \vec{x}^1_i)
    \end{equation}
    Starting from $t=2$, we estimate the expected position of each particle in the future frame as
    \begin{equation}
        \vec{p}_i^{t} =  \vec{x}^t_i +  \vec{v}^t_i \Delta t 
    \end{equation}
    We define the cost of association $\phi^n_{ij}$ to be the distance between particles $\vec{x}^{t+1}_j$ and the estimated position $\vec{p}_i^{t}$. A particle can be linked to the tracklet if the cost of linking is below a set threshold. The velocity corresponding to point $\vec{x}^{t+1}_j$ can be calculated as
    \begin{equation}
        \vec{v}^{t+1}_j = \frac{1}{\Delta t} (\vec{x}^{t+1}_j - \vec{x}^t_i)
    \end{equation}
    
    If multiple particles can be linked to the same tracklet, we stop the tracklet and start new ones. We set the threshold conservatively to minimize false linking. This results in shorter tracklets, which will be further connected in the re-tracking procedure described next. At last, the position and velocity of each point in a tracklet will be smoothed by a one dimensional Gaussian Filter \citep{mordant2004gaussian}. 
    
    \item \textit{Re-tracking}: Associate 3D tracklets to generate longer 3D tracks.
    All tracklets generated from the last stage are projected forward and backward in time using the positions and velocities at the endpoints \citep{xu2008tracking}. If distance between a forward projection of one tracklet is close to the backward projection of another tracklet, the two tracklets are joined. 
    When there are multiple possible matches, closeness of the velocity vectors is used to determine the best match. 
    In addition, we handle the transient disappearance and appearance of a particle from the field of view due to miss detection by extrapolation based on its previous motion history.
    At last, trajectories shorter than 10 frames are removed from the final set to avoid ghost trajectories.
\end{enumerate}

Generated tracks could be used to calculate motion priors of birds in the aviary, both of the collective as a whole as well as of the individuals. 

\subsection{Re-ID with the Bird15 dataset}
To form a dataset for bird re-identification, we exported images from stationary sequences. Images were passed through the bird detector and the sequence annotations (ground truth locations and identities of perched birds) were used to assign an identity with each detection. We exported tight crops from all available views, except when two or more birds occluded each other, in which case only the crop for the bird closest to the camera was exported for that view. To improve the spatial and pose diversity of exported crops, we partitioned the aviary into 3D bins (10 cm side length) and tracked the number of crops exported for each bird in each bin. For each bird, we exported crops every 10 frames until the bin for that bird and location had 10 images. Once the bin was filled, we continued to export crops, but only every 40 frames. We use this method to bias collection towards a diversity of locations generated by brief periods of perching as birds move throughout the aviary. All crops were resized to 256x256 pixels. Image filenames contain bird ID, camera view, sequence number, and frame number information following the Market1501 format \citep{Market1501}.

We split the dataset into training and test sets, composed of crops obtained from the first half and second half of each 15 minute segment, respectively. The training and test sets each contain 18,000 images. Birds were fairly evenly represented in both sets (mean $\pm$ std. training images per bird: $1225 \pm 531$, test images per bird: $1229 \pm 339$), with the exception of one female with Red+Yellow leg bands, which only had four examples in the training set and 620 in the test set. The number of examples from each of the top cameras was similar between training and test sets, and was consistently higher than the number of examples from the bottom cameras (as expected based on the lack of visibility of the perches). We randomly selected 7,500 training images to serve as a validation set.

We then trained an embedding network for bird identification on the Bird15 dataset. The network consists of a ResNet50 \citep{ResNet50} pre-trained on ImageNet, which takes in a 256 $\times$ 256 image and outputs a 2048 vector of re-ID features $f$, followed by a BNNeck \citep{Luo_2019_reid} and a classification head, which outputs identity logits $p$. The network was supervised using both triplet \citep{Weinberger2009TripletLoss} and cross-entropy identity losses and we used Adam and the FastReID codebase \citep{Luo_2019_reid} to optimize the model. We use the default FastReID baseline ``bag of tricks", except that we do not use horizontal flipping augmentation because bird identities depend on the ordering of the left/right leg band colors, which would be swapped upon reflection. During inference, we apply a softmax function to the logits $p$ to obtain a distribution over bird IDs for each image.

% \subsection{Sound detection and localization}

\section{Results and Experiments}\label{results_and_experiments}

\subsection{Short-term tracking of individual birds in cluttered scenes using WILD}
\begin{figure*}[hbt!]
\begin{center}
\includegraphics[width=\linewidth]{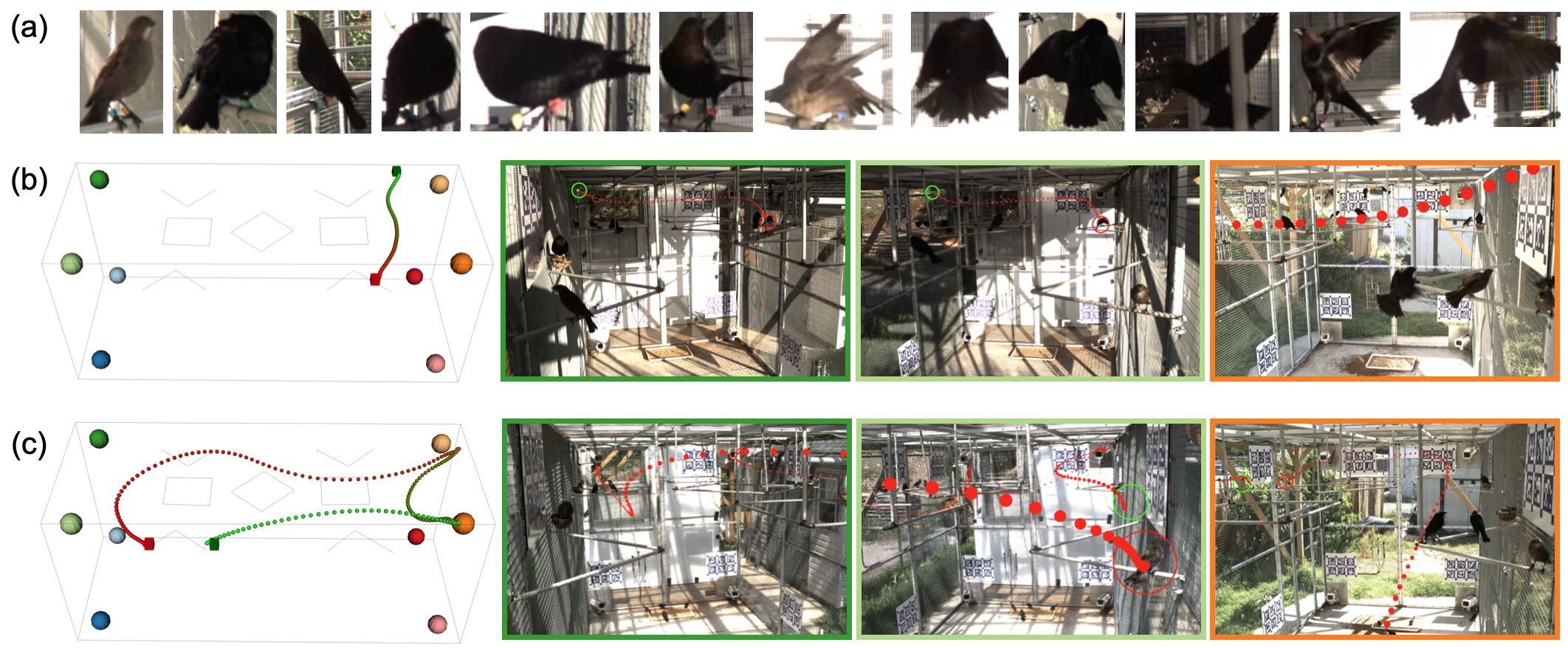}
\end{center}
\caption{\textbf{Qualitative tracking results.} (a) Examples of detected bird instances with variations in pose, shape, lighting, scale, occlusion, and motion blur. (b) Example of a successful short track (56 frames) followed by its 2D projections in 3 different views. Colors indicating the camera views are consistent with those in Figure \ref{fig:wild_dataset}. The green cube/circle is the start 3D/2D position and the red cube/circle is the end position. Dots in the 2D images are smaller/larger as the bird gets further away/closer. (c) Example of a successful long track (375 frames). During flight, the individual hops on the wall and briefly pauses for 1-2 seconds. Examples in (b) and (c) are from video segments drawn from different days, demonstrating variable time of day and lighting.}
\label{fig:tracking_results}
\end{figure*}

\begin{table*}[!htbp]
	\centering
	\caption{Quality of the trajectories retrieved by Stereo Matching method and Pointcloud Reconstruction method. AC0.1, AC0.3, AC0.5, and AC1.0 denote percent tracks land within 0.1m, 0.3m, 0.5m, and 1.0m of the ground truth end position, respectively.}
	\begin{tabular}{lcccccc}
		\toprule
			Method & Length (\# frames) & Segment Counts & AC0.1 & AC0.3 & AC0.5 & AC1.0  \\
		\midrule
			Stereo &  $\le 100$ & 741 & 0.17 & 0.34 & 0.41 & 0.52\\
			& $100 \sim 300$ & 186 & 0.10 & 0.20 & 0.27 & 0.47\\
			& $> 300$ & 25 & 0.04 & 0.08 & 0.20 & 0.28\\
		\midrule
			Pointcloud & $\le 100$ & 741 & 0.44 & 0.60 & 0.67 & 0.75\\
			& $100 \sim 300$ & 186 & 0.30 & 0.41 & 0.49 & 0.61\\
			& $> 300$ & 25 & 0.16 & 0.28 & 0.32 & 0.44\\
		
		\bottomrule
	\end{tabular}
	\label{tbl:tracking-result}
\end{table*}

\begin{table*}[!htbp]
	\centering
	\caption{Quality of the trajectories retrieved by our tracker assuming ``oracle" matching through ambiguities. }
	\begin{tabular}{cccccc}
		\toprule
			Length (\# frames) & Segment Counts & AC0.1 & AC0.3 & AC0.5 & AC1.0  \\
		\midrule
            $\le 100$ & 741 & 0.50 & 0.73 & 0.78 & 0.87\\
			$100 \sim 300$ & 186 & 0.45 & 0.62 & 0.67 & 0.76\\
			$> 300$ & 25 & 0.36 & 0.44 & 0.44 & 0.60\\
		
		\bottomrule
	\end{tabular}
	\label{tbl:oracle-matching-results}
\end{table*}

\textbf{Experiment.} 
We tested our tracker on the WILD dataset. 
Among the 952 motion segments we evaluated against, 741 segments have short sequences of $\le 100$ frames, 186 segments have $100 \sim 300$ frames, and 25 have rather long sequences of $\ge 300$ frames. 
For each motion segment, we provide the start and end locations of the target bird's head and tail points in 2D and 3D, as well as an iterator containing the sequence of synchronized multi-view frames. The task is to track the target bird and predict its 2D/3D position at the end of the sequence. 
The experiment was conducted as follows. 
We ran our multi-object tracker on the provided frame sequence and output a set of track hypotheses for all birds in the scene. At the start frame, we established correspondence between the target and the closest hypothesis based on 3D Euclidean distance, and at the end frame, we measured the 3D distance between the target's end location and the same hypothesis. All remaining hypothesis that were not associated with ground truth were ignored. 

We compared our Pointcloud Reconstruction based tracker with the Stereo Matching method introduced by \citet{ling2018stereomatching}. This method has been demonstrated to successfully resolve multi-view optical occlusions and improve tracking performance. The evaluation process for these two methods differs only in the point reconstruction stage, with the rest - detection and tracking - remaining the same (see Figure \ref{fig:tracking_pipeline}ABCD). One major difference of these two methods is the way they represent each target in 3D. Taking only the center of the detection mask/bounding box as input, the Stereo Matching method reconstructs the target as only one single point in space. The Pointcloud Reconstruction method, on the other hand, reconstructs the target as a dense cloud of points.

% Evaluation
\textbf{Evaluation metric.} 
The end position of the track hypothesis retrieved by our tracking pipeline, see Figure \ref{fig:tracking_results}, is compared with the ground truth end position. ``AC0.X'', the fraction of reconstructed hypotheses landing within 0.X meters of the ground truth, is reported in Table \ref{tbl:tracking-result}; its ideal value is equal to 100 percent. We chose this evaluation metric because distance based metrics were very sensitive to outliers. For example, samples that were not tracked successfully can land far away from the ground truth and end up dominating the average and inflating the standard deviation. We do not evaluate the result using the standard CLEAR MOT evaluation method of \cite{Bernardin2008}, because the MOT statistics are based on frame-by-frame annotations and the production of frame-by-frame 3D ground truth trajectories is currently severely limited by the amount of human effort and expertise required for manual annotation.

\begin{figure*}[hbt!]
\begin{center}
\includegraphics[width=\linewidth]{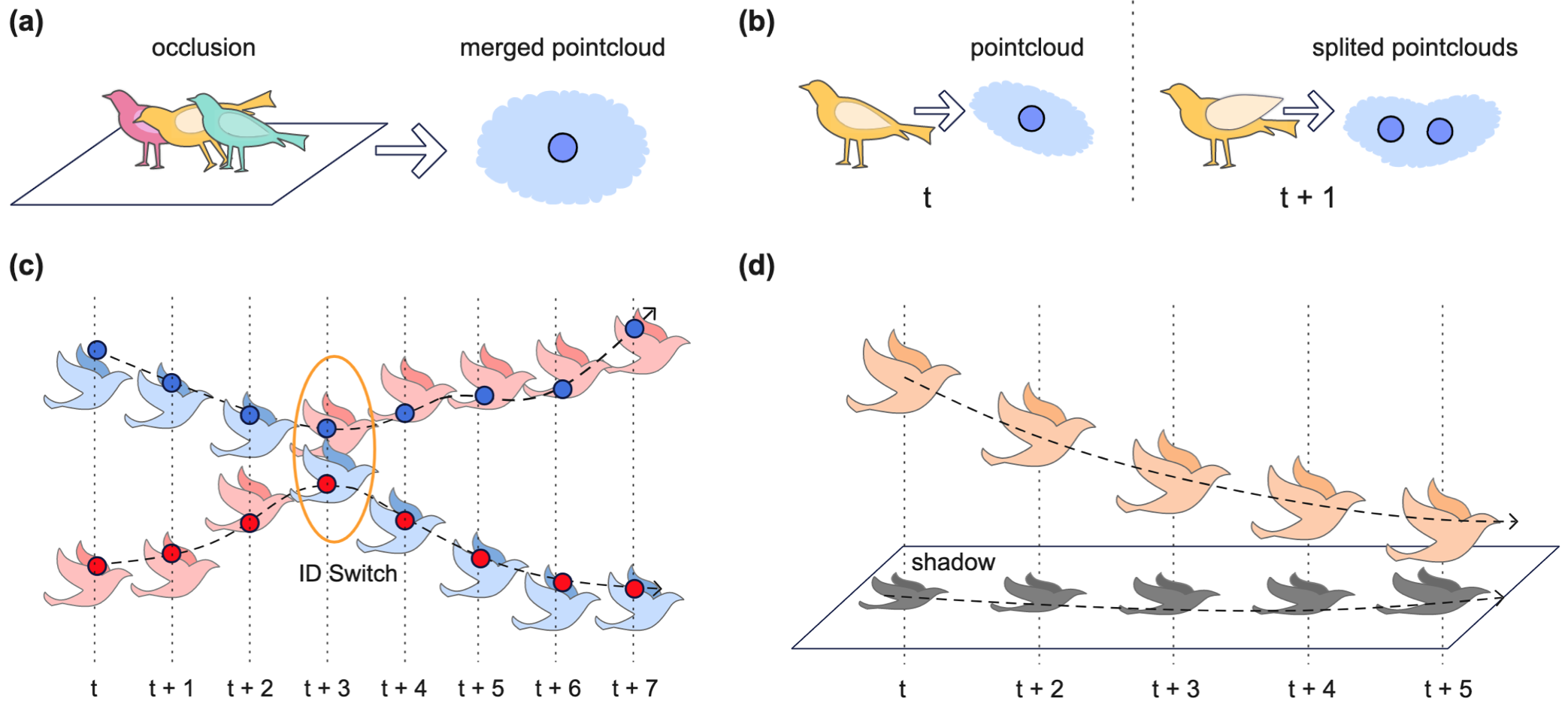}
\end{center}
\caption{\textbf{Failure cases.} (a) Inseparable pointcloud due to occlusions. (b) Merged/split clusters due to shape change of an individual at different instants of time, which could result in ghost trajectories. (c) Identity Switch. At first, the blue hypothesis is correctly tracking the ground truth blue bird. After a few frames, though, the blue bird and the red bird cross paths and blue hypothesis follows the wrong target. (d) Ghost trajectory resulting from false positive detections, eg. shadows of a bird.}
\label{fig:failure_cases}
\end{figure*}

% Results
\textbf{Result Analysis.}
We present qualitative results of the our tracker in Figure \ref{fig:tracking_results}. The quantitative results of both the Pointcloud Reconstruction method and the Stereo Matching method on the WILD dataset are reported in Table \ref{tbl:tracking-result}. The table shows that Pointcloud Reconstruction method outperforms the Stereo Matching method in every category. Video visualization shows that points reconstructed by Stereo Matching are more unstable than pointclouds, as the single-point representation is more sensitive to the quality of detections. A slight change of the detection (box size and shape) in the next frame will result in very different 2D location of the center and resulting reconstructed 3D points. 

As the tracking performance of the Stereo Matching method is significantly limited by the single-point representation, we restrict the following discussion to the Pointcloud Reconstruction method only. As Table \ref{tbl:tracking-result} shows, most tracks are either successful with low error (44\% of the short tracks land within 0.1m to the ground truth) or are not at all close (33\% of the short tracks land more than 0.5m from the ground truth). Increasing the threshold does not increase the overall accuracy very much. Table \ref{tbl:tracking-result} also shows that our tracker performs better on short segments than on the longer ones. To understand the influence of failures originating from ambiguities, we collected statistics of percent accuracy assuming ``oracle" matching through ambiguities. That is, we kept all possible matches during the re-tracking stage, and linked them to the tracking hypothesis to form a tree structure. We counted a hypothesis as a success as long as one of it's leaf nodes landed within the threshold of the ground truth. Statistics are reported in Table \ref{tbl:oracle-matching-results}. As the table shows, accuracy of the longer tracks has increased notably, indicating ambiguities are an important source of failure. This problem could be aided by re-ID or visual features as discussed in the next section.

Assuming failures are solely due to accumulating errors ambiguities or missed detections are encountered, if 44\% of tracks are successful for 100 frames, then we can expect only 19\% of tracks to survive to 200 frames and 9\% to survive to 300 frames. Because the performance is better than this expectation, it is possible that the tracker is struggling elsewhere. For example, during initialization, there might be no track available to assign to the target start, or the wrong track could be assigned to the target start. A discussion of the failure cases is provided in the next paragraph.

\textbf{Failure cases catalogue.}
Our tracker produced many plausible results but also many failure cases, shown in Figure \ref{fig:failure_cases}. To better understand the nature of the complexity of the WILD dataset, we manually examined 20 failure cases by looking into the outputs (detections, pointclouds, tracklets) produced in each stage of the pipeline frame by frame. We found that the tracker struggles in the following cases:
\begin{enumerate}
    \item Missed detections: extreme poses and occlusions from poles and other individuals in the aviary occasionally cause the detector to fail.
    \item False positive detections: shadows of birds, for example, create ghost pointclouds and ghost trajectories (Figure \ref{fig:failure_cases}d).
    \item An inseparable pointcloud due to occlusions (Figure \ref{fig:failure_cases}a): multiple targets in close 3D proximity can occlude each other in all camera views. They then become reconstructed as one pointcloud as a whole and share one track. 
    \item Merged and split pointclouds: when individuals change shape or size (Figure \ref{fig:failure_cases}b), pointclouds can split into two or more clusters. During flight, the appearance of a bird changes dramatically in a very short period of time (Figure \ref{fig:tracking_results}a), which results in differently shaped clouds of points. In many cases, points representing one bird are grouped into multiple clusters (Figure \ref{fig:failure_cases}b), which introduces unstable and unpredictable ghost pointclouds. Such instability increases the difficulty of tracking.
    \item Identity switches: true identities of different hypotheses can become switched, particularly if two individuals remain directly next to each other for several seconds (Figure \ref{fig:failure_cases}c).
\end{enumerate}

\subsection{Bird re-identification}
\begin{figure*}[hbt!]
\begin{center}
\includegraphics[width=0.85\linewidth]{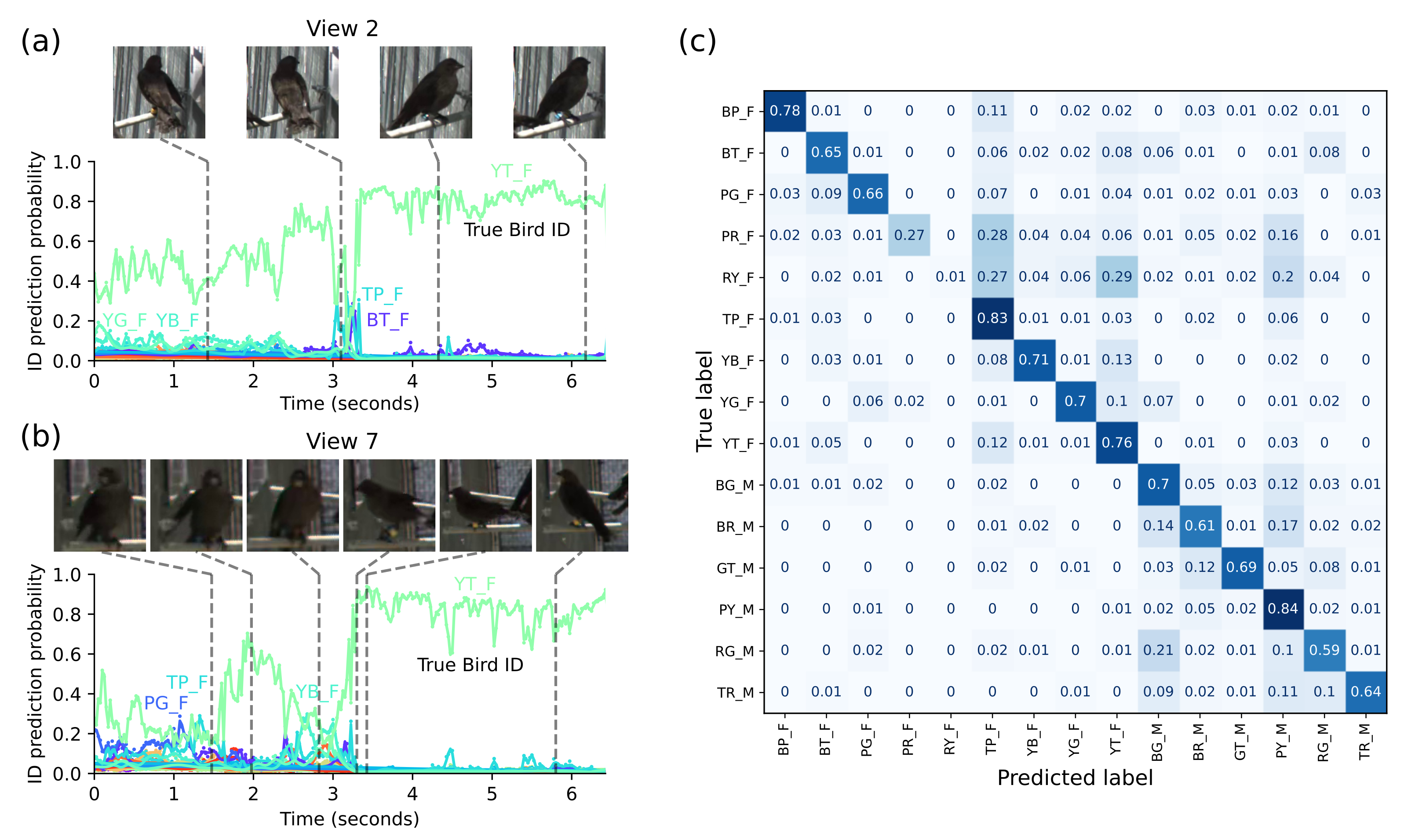}
\end{center}
\caption{\textbf{Bird re-identification.} We use a ResNet50 network supervised with triplet and ID losses to predict the identity of perched birds. In an example from the Bird15 test, a female with Yellow+Teal leg bands is visible from views 2 (a) and 7 (b). From view 2 (a) only its left leg band is initially visible, but the network has learned other features (such as tail shape, or background features if the bird is in a repeatedly used location) that allow it to correctly predict the identity. When no bands are visible (second image from the left in a), the confidence decreases. Once both bands are visible (third and forth images) confidence increases again. From another view (b), both bands are visible, but are in a shadow and some initial color distortion causes the network to incorrectly predict Pink+Green, Teal+Pink, and Yellow+Blue, albeit with low confidence. As the bird reorients to face the other direction, both bands become visible with better lighting and confidence increases. A normalized confusion matrix (c) shows most birds are correctly identified 60--80\% of the time in the test set. Increasing the detection confidence threshold from 0 to 0.8 improves accuracy from 0.68 to 0.97 while still correctly identifying 52\% of the examples in the Bird15 test set.}
\label{fig:bird_reid}
\end{figure*}
We evaluate the performance of the re-ID network using the Bird15 test set, which we constructed using the ground truth locations of perched birds. Overall, the network correctly identified 68\% of examples in the test set and most individuals are identified correctly 60--80\% of the time (Figure \ref{fig:bird_reid}c). Instead of returning whichever bird corresponds to the highest probability (even if it is very low), setting a detection confidence threshold to 0.8 increases the accuracy to 0.97 while correctly predicting 52\% of samples in the test set. Most confusion appears to be within females and within males separately, with relatively low confusion between males and females. Unless lighting is very poor, males can usually be distinguished from females by their darker color.

When deployed on crop sequences from tracked birds (Figure \ref{fig:bird_reid}a,b), probability trajectories over time reveal interesting patterns of the re-ID network. From camera view 2 (Figure \ref{fig:bird_reid}a), the network predicts the correct identity despite only being able to see one band (three other female birds have yellow bands). When both bands are hidden, however, the network becomes less confident. Interestingly, these observations suggest that the network has learned to rely on the bands, but that it has also learned to rely on additional features such as slight variations in bird color or patterning, or perhaps features of the background behind the favorite perch locations for each bird. This hypothesis could be tested by training on a masked dataset, where the network receives only pixels corresponding to the bird and no pixels from the background. Improving the diversity of perch positions by collecting additional annotations throughout the breeding season may also help improve the robustness of the bird re-ID pipeline.

\subsection{Social network analysis} \label{socialnetworkanalysis}
Using our dataset we analyzed the birds’ social network and investigated how birds’ behavior depends on social context. In addition to human labeled song annotations, we also added “approach”, “stay”, “leave”, and “sing to” interactions using the start and endpoints of the stationary sequences. Whenever a bird flew to a location within an interaction distance (0.5 meters) of another, we added a “b1 approached b2” annotation. Whenever a bird was within the interaction distance of another and flew away we added a “b1 left b2” annotation. Whenever a male sang, we added “b1 sang to b2” annotations for all birds within the interaction distance. Finally whenever a bird was approached, if it did not leave within one second, we add a “b1 stayed with b2” annotation \citep{Anderson2021}. After collecting the interactions between all pairs of birds, we grouped interactions depending on social context factors, such as those belonging to male-male interactions, or those between a pair-bonded or non-pair-bonded male and female. We defined a pair bond between a male and a female whenever the female received more than 50\% of her total song interactions from that specific male \citep{Anderson2021}. From the sets of interactions, we constructed transition ethograms and inspected how the probabilities of interaction transitions changed with social context. We focus our analyses on two 15 minute segments with song annotations from mid May.

\begin{figure}[hbt!]
\begin{center}
\includegraphics[width=0.98\linewidth]{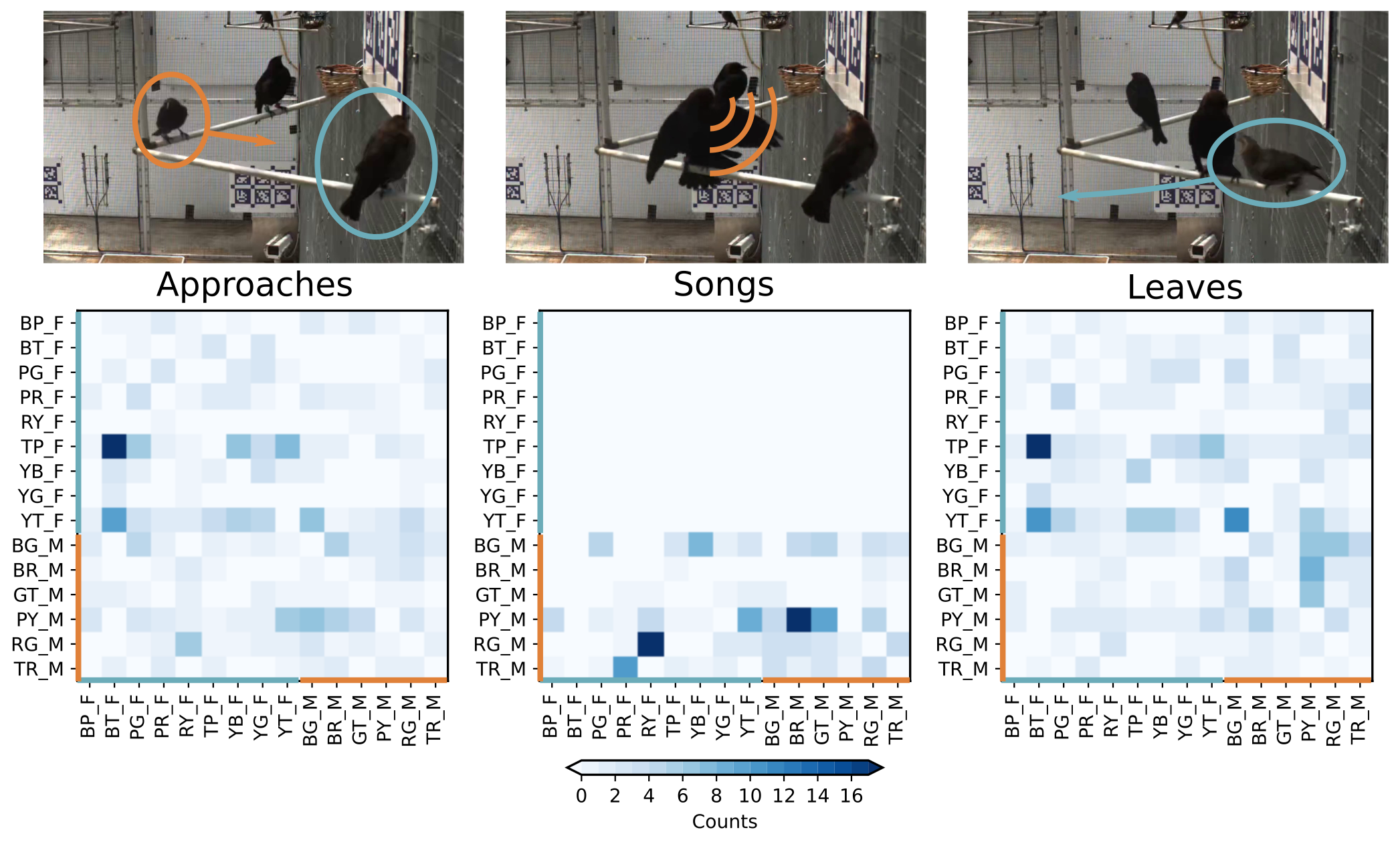}
\end{center}
\caption{\textbf{Pairwise interactions.} Approaches, songs, and leave interactions occur frequently between individuals in the aviary. Each matrix shows the frequency of interactions for each pair of individuals. The bird performing the action is shown on the left axis (the approaching, singing, or leaving bird) and the target or recipient of the action is shown along the bottom axis (the approached, receiving, or remaining bird). Orange indicates males and blue indicates females. Approaches and leaves show relative movement between individuals and reveal differences in activity levels and dominance (see section \ref{socialnetworkanalysis}). We also observed six pair bonds between males and females, which are defined whenever a female receives more than 50\% of songs from a single male \citep{Anderson2021}.}
\label{fig:bird_interactions}
\end{figure}

From the patterns of approaches and leaves, we observed differences in the overall activity levels of individuals (Figure \ref{fig:bird_interactions}). Two females, Teal+Pink and Yellow+Teal, repeatedly flew back and forth among two or more perches, one of which was within the interaction distance of where Blue+Teal was perched. The approach and leave interaction data among males revealed that male Pink+Yellow frequently approaches Blue+Green, Blue+Red, and Green+Teal males (darker \verb|PY_M| row in the approaches matrix), and at the same time, these three males frequently fly away from Pink+Yellow (darker \verb|PY_M| column in the leaves matrix). These patterns clearly indicate that Pink+Yellow is dominant over these males.

\begin{figure*}[hbt!]
\begin{center}
\includegraphics[width=0.92\linewidth]{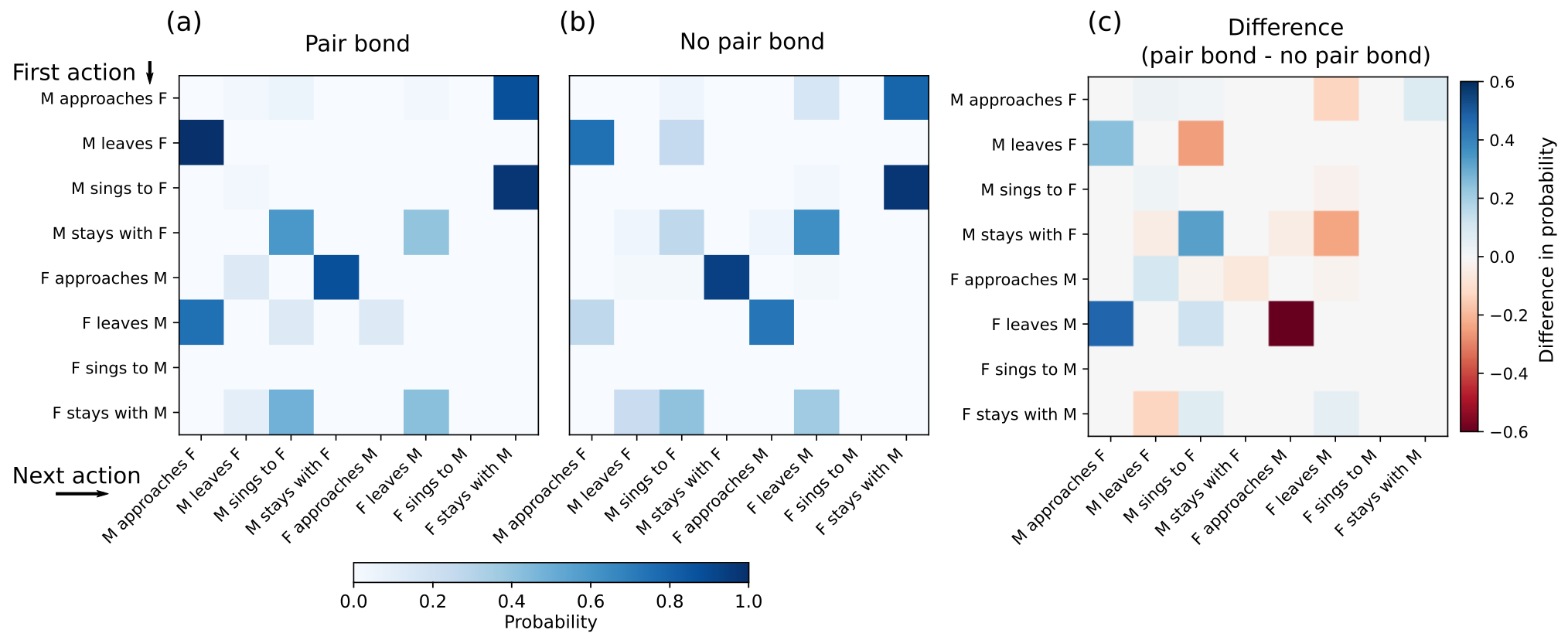}
\end{center}
\caption{\textbf{Interaction sequences.} Interaction transition probabilities differ between pair-bonded (a, n = 163 transitions) and non-pair-bonded (b, n = 187 transitions) males and females. For a given row, filled-in cells show interactions that occurred next based on their frequency in the dataset. Counts are normalized within rows and darker blue shows greater probability. (c) The difference in transition probabilities for bonded pairs relative to non-bonded pairs. Darker blue indicates a transition is more likely for a bonded pair than for a non-bonded pair; darker red indicates a transition is more likely for a non-bonded pair than a bonded pair. Transition probabilities reveal that pair-bonded females are generally more receptive to approaches by their pair bond male than by other males and that pair-bonded males are more likely to follow females with which they have formed a pair bond.}
\label{fig:interaction_sequences}
\end{figure*}

From the song interaction data, we observed six pair bonds between males and females. Both Blue+Pink and Yellow+Teal females were pair bonded with the Pink+Yellow male. Similarly, Pink+Green and Yellow+Blue females were bonded with the Blue+Green male. Red+Yellow and Pink+Red females were bonded with Red+Green and Teal+Red males, respectively. Based on these pair bonds, we split the set of interaction transitions into pair bond and non-pair bond groups (Figure \ref{fig:interaction_sequences}). Inspecting the differences in transition probabilities of pair-bonded birds relative to non-pair-bonded birds (Figure \ref{fig:interaction_sequences}c) reveals that females are more likely to leave when approached by non pair bond males than when approached by their pair bond male. When a female stays with its pair bond male, the male is more likely to sing to her and less likely to leave than when a female stays near a non pair bond male. When a female leaves her pair bond male, the male is more likely to follow and approach her again, than when a female leaves a non-pair bond male.

It will be interesting to analyze how patterns of interaction vary throughout time of day and over the breeding season. For example, in one of the annotated 15 minute segments in April, males were actively singing for nearly the entire period, but we recorded very few flight sequences, leaves, and approaches because most birds remained on their perches. Without many more periods of observation, it will remain unclear whether such differences in interaction patterns are a normal part of social network formation, or whether they can be explained by other environmental variables such as time of day, temperature, and weather. 

\begin{figure*}[hbt!]
\begin{center}
\includegraphics[width=0.92\linewidth]{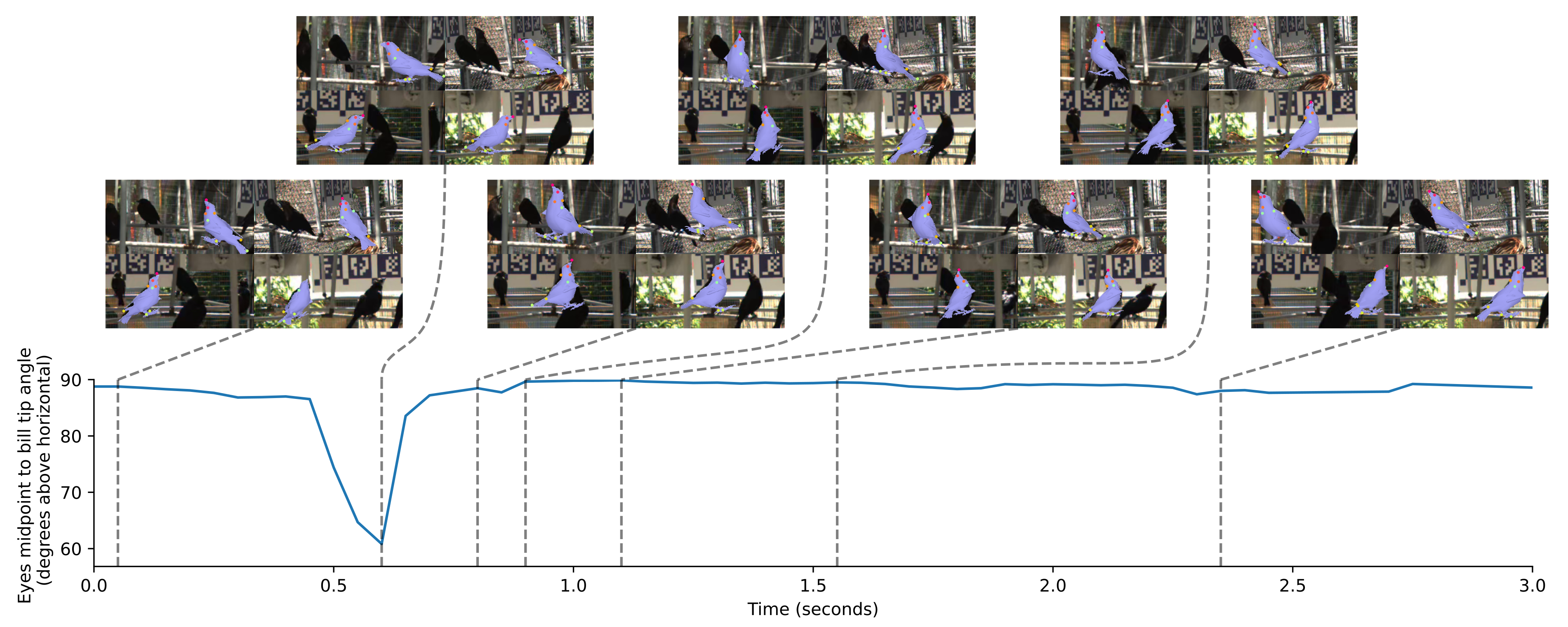}
\end{center}
\caption{\textbf{Pose trajectories.} Behaviors extracted from pose trajectories can reveal fine-grained interactions such as head-up aggressive displays by males. In every other frame, a three dimensional parameterized mesh \citep{badger2020} is fit to multi-view anatomical keypoints. In this example, the angle between horizontal and the vector from the midpoint between the eyes to the bill tip (visualized in the plot) captures this behavior well.}
\label{fig:pose_trajectory}
\end{figure*}

Finally, we anticipate that estimating the pose and shape of individuals in the aviary \citep{badger2020} will allow us to incorporate more fine-grained behaviors and interactions, such as the head-up display shown in Figure \ref{fig:pose_trajectory}.

\section{Conclusion}

In this work we develop a system for capturing the behavioral interactions of a group of 15 songbirds. Although we found that our pointcloud reconstruction method performed better than a stereo matching method, there is still much room for performance improvements on our difficult multi-view multi-animal Where'd It LanD (WILD) dataset. We introduce several complexities that arise when studying animals that maneuver and interact in three dimensions. Tracking many individuals across multiple sensors is a challenging task with points of failure. The relative lack of flying birds in our detection dataset (birds spend most of their time sitting perched) hindered our object detection pipeline and lead us to add the additional complexity of a motion detector. Replacing this motion detector with a neural network designed specifically for detecting objects in motion could significantly improve our pipeline by reducing the number of false positive detections (and ensuing ghost trajectories and tracking failures) generated by background motion. We also found that birds occluded each other much more than expected because the perches were positioned only slightly below plane of the top cameras. We plan to improve the layout of the aviary in order to reduce such occlusions. We also highlight the need for additional work that integrates detection, tracking, re-ID, and pose estimation pipelines without relying on extensively annotated tracking datasets, which become prohibitively expensive to create in multi-view multi-animal settings. Using our system and dataset of ground-truth identities, we developed a re-ID pipeline, extracted detailed ethograms for all birds in the aviary, and demonstrated that the presence of a pair bond changes the interaction dynamics between males and females.

\backmatter

\section*{Acknowledgements}
We are grateful for the help of Henry Korpi, Ana Alonso, Greg Forkin, and Marcelina Martynek for their helpful discussion and many contributions to annotations in the dataset.

\section*{Declarations}
\subsection*{Competing interests}
The authors declare no competing or conflicts of interest. 

\subsection*{Ethics approval}
The aviary and cowbird data collection were approved by the University of Pennsylvania Institutional Animal Care and Use Committee.

\subsection*{Funding} We gratefully acknowledge support through the following grants: National Science Foundation IOS-1557499, National Science Foundation MRI 1626008, National Science Foundation NCS-FO 2124355.

\subsection*{Data and code availability}
Data and code will be made publicly available via Google Drive and GitHub.

\subsection*{Authors' contributions}
M.S. and K.D. conceived of the study.
A.P., B.P., and M.S. constructed the aviary and collected the data.
M.B., S.X., Y.W., and K.D. designed the tracking approaches and dataset. 
M.B., S.X., and Y.W. developed the tracking and re-ID pipelines. 
%M.B., A.P, A.A., G.F. and M.M developed sound detection and localization. 
%M.B., A.A., G.F. collected the dataset. 
M.B. and A.P. prepared the dataset.
M.B., S.X., and Y.W. performed the experiments and created the figures. 
M.B. and S.X. wrote the first draft. 
M.B., S.X., Y.W., M.S. and K.D. edited the paper for submission.

% \item Conflict of interest/Competing interests (check journal-specific guidelines for which heading to use)
% \item Ethics approval 
% \item Consent to participate
% \item Consent for publication
% \item Availability of data and materials
% \item Authors' contributions

% \begin{appendices}

% \section{Section title of first appendix}\label{secA1}

% An appendix contains supplementary information that is not an essential part of the text itself but which may be helpful in providing a more comprehensive understanding of the research problem or it is information that is too cumbersome to be included in the body of the paper.

% %%=============================================%%
% %% For submissions to Nature Portfolio Journals %%
% %% please use the heading ``Extended Data''.   %%
% %%=============================================%%

% %%=============================================================%%
% %% Sample for another appendix section			       %%
% %%=============================================================%%

% %% \section{Example of another appendix section}\label{secA2}%
% %% Appendices may be used for helpful, supporting or essential material that would otherwise 
% %% clutter, break up or be distracting to the text. Appendices can consist of sections, figures, 
% %% tables and equations etc.

% \end{appendices}

\clearpage

%\bibliographystyle{apalike}
\bibliography{sn-bibliography}% common bib file


\end{document}